\documentclass[11pt]{article}

\usepackage[final]{acl}

\usepackage{times}
\usepackage{latexsym}

\usepackage[T1]{fontenc}

\usepackage[utf8]{inputenc}

\usepackage{microtype}

\usepackage{inconsolata}

\usepackage{graphicx}
\usepackage{amsmath}
\usepackage{booktabs}

\title{MAS-Bai: Linguistically Informed Adaptive Multi-Agent Translation for Bowen-Script Bai}
\hypersetup{
  pdftitle={MAS-Bai: Linguistically Informed Adaptive Multi-Agent Translation for Bowen-Script Bai},
  pdfauthor={Jingting Zheng, Hong Ren, Yuehan Cui, Shaolin Zhu, Deyi Xiong}
}

\author{
Jingting Zheng\thanks{Equal contribution.}
  \, Hong Ren\footnotemark[1]
  \, Yuehan Cui, Shaolin Zhu, Deyi Xiong\thanks{Corresponding author.}\\
TJUNLP Lab, School of Computer Science and Technology, Tianjin University, China\\
\texttt{\{zhengjingting, rhong, 3024244124, zhushaolin, dyxiong\}@tju.edu.cn}
}

\begin{document}
\maketitle
\begin{abstract}
Large Language Models (LLMs) remain unreliable for extremely low-resource languages whose writing systems conflict with high-resource pretraining priors. We study Bai--Chinese translation in the Bowen script, where Chinese characters are frequently borrowed phonetically rather than semantically, causing severe literal-translation hallucinations. We propose \textbf{MAS-Bai}, a linguistically informed adaptive multi-agent framework for Bowen-script Bai--Chinese translation. MAS-Bai routes each input through either a lightweight retrieval--translation path or a deeper philological path with etymological and syntactic agents, and uses cycle-consistency with a lightweight reverse translator as an inference-time reliability signal. We construct a 6,536-sentence Bai--Chinese corpus and evaluate on a 1,536-sentence test set. MAS-Bai achieves 16.23 BLEU and 62.15 COMET, improving over Standard RAG by 8.11 BLEU and 6.85 COMET. Forward LoRA, expert ablations and controlled routing experiments distinguish the roles of supervised adaptation, Bai-specific expertise and conditional routing. The largest gains occur on Bowen phonetic loans and divergent Bai word order, while human evaluation and reverse-verifier analyses support the automatic results.
\end{abstract}

\section{Introduction}

Large Language Models (LLMs) have fundamentally reshaped Machine Translation (MT), demonstrating strong capabilities for high-resource languages through in-context learning (ICL) and instruction tuning \citep{achiam2023gpt,team2023gemini,dubey2024llama,liu2024deepseek}. Despite these advances, extending this success to extremely low-resource and linguistically complex languages remains a formidable bottleneck \citep{chen2021machine,zhu2024landermt}. A representative example is Bai, a Sino-Tibetan language spoken in Yunnan, China. Unlike languages with standardized alphabetic scripts, Bai traditionally employs the Bowen script, a logographic system in which Chinese characters are extensively borrowed or adapted on the basis of phonetic similarity and locally established writing conventions \citep{yeon2008no}. Consequently, a character in a Bowen-script Bai sentence may be graphically identical to a Chinese character while expressing a different meaning or grammatical function.

Conventional Neural Machine Translation (NMT) and standard LLM fine-tuning \citep{touvron2023llama,zhu2024multilingual} are poorly matched to this setting for three reasons. First, script opacity induces multilingual interference: pretrained LLMs often map Chinese-looking characters to standard Mandarin meanings, which can override the Bai context. Second, the limited training data are insufficient to counter these priors. Third, fixed inference workflows apply the same processing pipeline to inputs ranging from transparent dialogue-like utterances to opaque phonetic-loan cases.

Recent computational work on Bai has begun to examine its phonology, including IPA-to-Pinyin correspondence \citep{yang2025study} and spoken-only translation through phonetic transcription \citep{chen2025can}. However, machine translation of the historical logographic Bowen script remains largely unaddressed. Bai texts vary widely in complexity, from straightforward daily speech to highly metaphorical folk songs \citep{li2015bai}. General reasoning methods such as Chain-of-Thought (CoT) prompting \citep{wei2022chain} decompose tasks, but they do not incorporate the etymological strategies needed for deciphering Bowen-script Bai.

To address these challenges, we propose \textbf{MAS-Bai}, an inference-time adaptive multi-agent framework for Bowen-script Bai--Chinese translation. MAS-Bai uses a Meta-Agent to detect explicit Bai/Bowen cues, such as ambiguous phonetic-loan characters and divergent syntactic patterns, and routes each input to either a lightweight retrieval--translation path or a deeper philological path. For complex inputs, the framework activates specialized agents for character decipherment, structural reordering, and retrieval from our 6,536-pair corpus. During inference, a cycle-consistency loop with a lightweight reverse translator evaluates draft translations and refines the selected topology when the draft appears inconsistent with the original Bai input. The expert definitions and routing cues are tailored to Bai. More broadly, the framework uses explicit linguistic evidence to select an inference pathway at test time.

Our contributions are summarized as follows.
\begin{itemize}
\setlength{\itemsep}{0pt}
\setlength{\parsep}{0pt}
\item We introduce \textbf{MAS-Bai}, a linguistically informed adaptive multi-agent framework for Bowen-script Bai--Chinese translation that combines Bai-specific philological experts with inference-time conditional routing.

\item We construct a 6,536-sentence Bai--Chinese corpus and annotate the 1,536-sentence test set by register and linguistic phenomena, providing the basis for stratified evaluation of script opacity and divergent syntactic patterns.

\item We show that MAS-Bai outperforms strong zero-shot, ICL, Standard RAG, Forward LoRA and Static MAS baselines. Expert ablations and controlled routing comparisons assess the contributions of Bai-specific linguistic knowledge and pathway selection, supplemented by stratified, human and cycle-consistency analyses.

\end{itemize}

\begin{figure*}[t]
  \centering
  \includegraphics[width=0.98\textwidth]{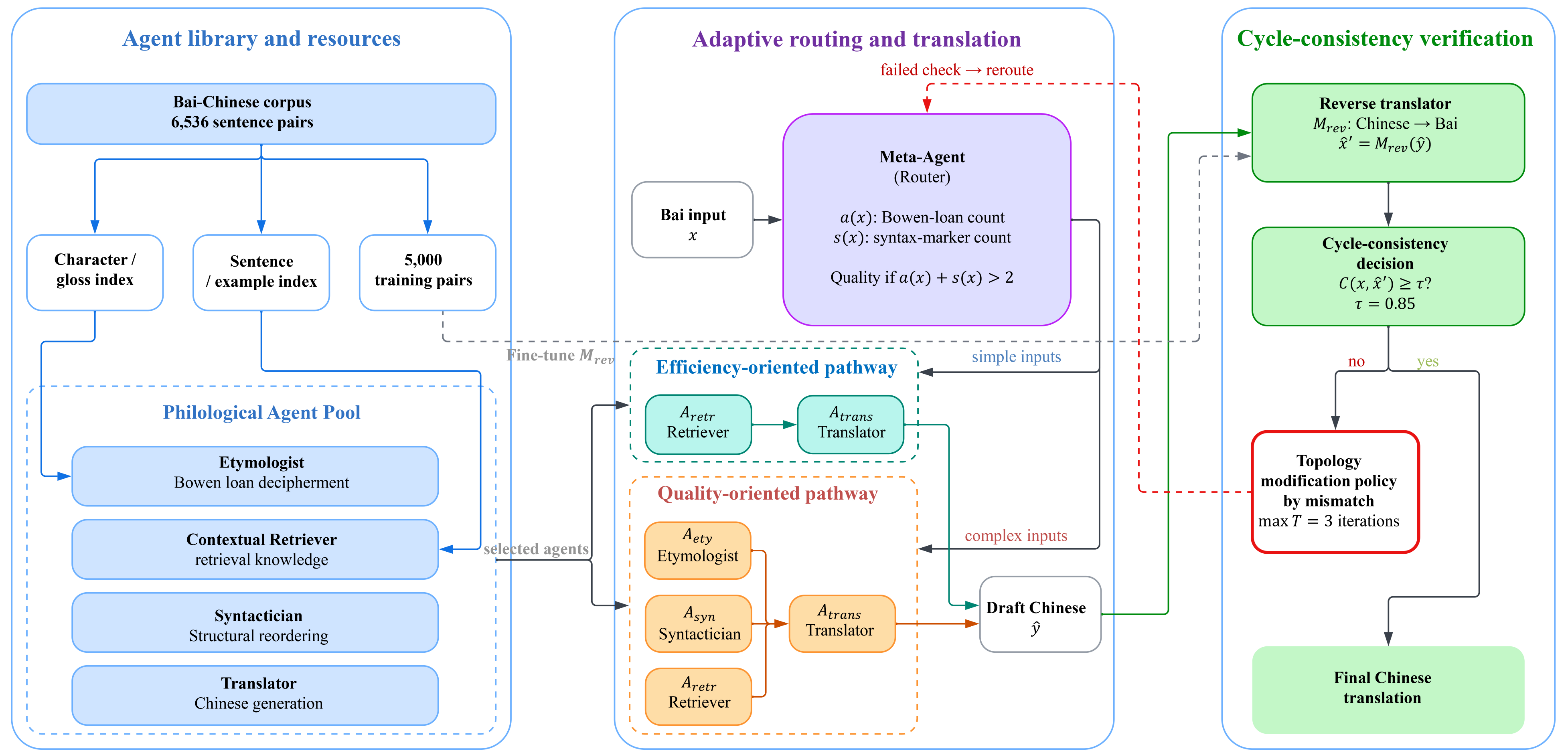}
  \caption{
Architecture of MAS-Bai. The Meta-Agent routes inputs by Bowen-loan and syntax-marker counts to either a retrieval--translation path or a philological expert path. Cycle-consistency failures trigger topology modification for up to $T=3$ iterations.
}
  \label{fig:architecture}
\end{figure*}

\section{Related Work}

MAS-Bai draws on three areas of research: low-resource machine translation, multi-agent workflow design and LLM agents for translation. We review these areas to clarify the main distinction of our approach: explicit Bai/Bowen linguistic evidence guides the selection of the inference workflow for each input.

\paragraph{LLMs for Low-Resource Machine Translation.}
Massively multilingual models such as NLLB \citep{costa2022no} and Aya \citep{ustun2024aya} expand language coverage, but their performance often remains limited on extremely low-resource languages because of the curse of multilinguality \citep{conneau2020unsupervised} and negative transfer from dominant scripts. Parameter-efficient methods such as LoRA \citep{hu2022lora} adapt generalist models without updating all backbone parameters. Language-aware and modular MT approaches further distinguish between shared and language-specific model capacity through gating, modularization or parameter routing \citep{sun2022languagebranch,li2023mmnmt,zhu2024landermt,dong2025mlaslora}. Whereas these methods primarily adapt model parameters or modules, MAS-Bai operates at the workflow level by selecting the inference pathway according to sentence-level philological complexity.

\paragraph{Multi-Agent Systems and Workflow Design.}
Multi-agent systems decompose complex tasks into collaborative workflows, and early frameworks relied mainly on human-defined static topologies \citep{li2023camel,chen2024agentverse}. More recent work treats workflow design itself as an optimization problem. For example, MAS-ZERO \citep{ke2025mas} and AFlow \citep{zhang2024aflow} search for effective operators or agent graphs during training or validation. Self-evolving agent frameworks extend this line of work by refining workflows through iterative feedback after their initial design \citep{zhao2025absolute,huang2025r}. This line of work treats the agent graph as adaptable, but the adaptation is usually driven by task-level performance signals or generic search objectives.
MAS-Bai instead grounds inference-time topology selection in explicit sentence-level Bai/Bowen cues, using phonetic-loan and syntax markers to route each input.

\paragraph{LLM Agents for Translation.}
Recent translation-agent systems have explored memory, reflection, long-reasoning and online adaptation workflows for MT. DelTA uses multi-level memory for online document-level translation, improving document consistency through proper-noun records, bilingual summaries, and short- and long-term memory \citep{wang2025delta}. DUAL-REFLECT improves LLM translation through dual-learning feedback and reflective revision \citep{chen2024dualreflect}. DRT introduces long chain-of-thought translation for literary sentences involving similes and metaphors \citep{wang2025drt}. AdaDPI uses linguistic uncertainty to trigger online parameter internalization for document-level translation \citep{ren2026adadpi}. These systems are especially relevant to MAS-Bai because they formulate translation as a multi-stage inference process. In MAS-Bai, explicit Bai/Bowen cues guide the selection of external philological agent topologies, enabling adaptation at the workflow level.

\section{Bai--Chinese Parallel Corpus}
\label{sec:dataset}

We construct a sentence-level Bai--Chinese parallel corpus for Bowen-script Bai translation, paired with Modern Chinese references. The corpus is compiled from two authorized Bowen-script documentation sources: an officially published \textit{Dabenqv} collection \citep{000} and a community-held handwritten Bowen-script manuscript. After digitization, translation, sentence-level alignment and quality control, the final corpus contains 6,536 Bai--Chinese sentence pairs. Detailed construction procedures, source-level counts and token statistics are provided in Appendix~\ref{app:corpus_construction}.

To analyze performance across different registers and linguistic phenomena, we annotate the 1,536-sentence test set along two axes. The first axis is sentence-level register, including dialogue-like utterances, formal narrative passages and lyric or verse lines. The second axis marks Bai-specific linguistic phenomena, including translation-critical Bowen phonetic loans and Bai word-order patterns that diverge from standard Chinese. These annotations are used in Section~\ref{sec:analysis_phenomena}.

\section{Inference-Time Adaptive Multi-Agent Collaboration Framework}
\label{sec:framework}

Figure~\ref{fig:architecture} shows the overall architecture of MAS-Bai, which consists of three components: a philological agent pool, a Meta-Agent that conditionally routes inputs and a cycle-consistency self-evolution loop.

\subsection{Philological Agent Pool}
\label{sec:agent_pool}

To address the linguistic heterogeneity of Bowen-script Bai, MAS-Bai decomposes translation into specialized capabilities: character decipherment, syntactic reordering, contextual retrieval and Chinese generation. The curated 6,536-pair Bai--Chinese corpus described in Section~\ref{sec:dataset} serves as the retrieval and grounding resource for these agents. The agent roles are defined as follows.

\paragraph{Etymologist.} 
The Etymologist targets script opacity, especially cases where Chinese characters are used in Bowen script as phonetic loans whose Bai usage is unrelated to their standard Chinese semantics. We construct a character-level index from the parallel corpus. When activated, the Etymologist identifies potentially ambiguous Bowen characters in the input, retrieves usage examples from the Knowledge Base, and produces philological glosses. These glosses distinguish whether a target character functions as a semantic morpheme or as a phonetic loan in the current Bai context.

\paragraph{Syntactician.}
The Syntactician targets structural divergence between Bai and Modern Chinese. Some Bai constructions follow word-order patterns that differ from standard Chinese, including SOV-like structures and postpositional patterns. Direct generation can therefore produce unnatural or erroneous Chinese word order. The Syntactician analyzes the input structure and produces a reordered intermediate representation that approximates target-side Chinese word order. This agent does not translate content directly but reduces the structural burden on the final generation step.

\paragraph{Contextual Retriever.}
The Contextual Retriever implements retrieval-augmented generation over the Bai--Chinese Knowledge Base. Given the sensitivity of LLM-based MT to the relevance and noise of in-context demonstrations \citep{zhu2024robusticl}, it retrieves the top-$k$ semantically similar sentence pairs as dynamic few-shot examples, providing local evidence for lexical choice, register and phrasing.

\paragraph{Translator.}
The Translator is the Chinese generation agent. It synthesizes the available upstream evidence, including character glosses from the Etymologist, structural cues from the Syntactician, and retrieved examples from the Contextual Retriever, to produce a draft Chinese translation $\hat{y}$. The draft is then passed to the cycle-consistency loop before final acceptance.

\subsection{Conditional Meta-Architecting Mechanism}
\label{sec:meta_architecting}

MAS-Bai uses conditional meta-architecting to balance translation quality and computational cost. Before translation, the Meta-Agent performs a lightweight feature check, and uses the resulting cues to select either the efficiency-oriented or quality-oriented pathway.

\paragraph{Complexity features.}
Given a Bai input sentence $x$, the Meta-Agent extracts two feature counts. Let $a(x)$ denote the number of detected characters drawn from a curated list of high-frequency ambiguous Bowen phonetic loans, and let $s(x)$ denote the number of detected syntactic markers that diverge from standard Chinese word order, including postpositional particles and SOV-like patterns. The routing rule is deterministic:

\begin{equation}
r(x)=
\begin{cases}
\textsc{Quality}, & a(x)+s(x) > 2,\\
\textsc{Efficiency}, & \text{otherwise}.
\end{cases}
\end{equation}

The threshold was selected using a 100-sentence development sample and then fixed for all test-set experiments. The prompt-based Meta-Agent applies this deterministic rule to select the inference pathway for each input.

\paragraph{Initial topology.}
For inputs routed to the efficiency-oriented pathway, MAS-Bai uses a lightweight retrieval--translation graph:
\[
x \rightarrow A_{\mathrm{retr}} \rightarrow A_{\mathrm{trans}} \rightarrow \hat{y}.
\]
For inputs routed to the quality-oriented pathway, MAS-Bai activates the full philological graph:
\[
x \rightarrow A_{\mathrm{ety}}, A_{\mathrm{syn}}, A_{\mathrm{retr}} \rightarrow A_{\mathrm{trans}} \rightarrow \hat{y}.
\]
The former avoids unnecessary expert calls for transparent or dialogue-like inputs, whereas the latter adds etymological and syntactic analysis for opaque Bowen-script inputs or cases with Bai word-order patterns that diverge from standard Chinese.

\subsection{Inference-Time Self-Evolution via Cycle-Consistency}
\label{sec:self_evolution}

We use cycle-consistency as an inference-time reliability signal. The reverse translator $M_{\mathrm{rev}}$ is a Qwen-2.5-7B model fine-tuned to map Chinese outputs back into Bai. Given a draft Chinese translation $\hat{y}$, the system generates a reconstructed Bai sentence:
\[
\hat{x}' = M_{\mathrm{rev}}(\hat{y}).
\]

The verifier compares $\hat{x}'$ with the original Bai input $x$ using a composite consistency score $C(x,\hat{x}')$. This score combines two signals: sentence-level embedding similarity that captures coarse semantic consistency, and character-level overlap over Bai/Bowen markers that checks whether translation-critical source markers are preserved through back-translation. We fix the acceptance threshold to $\tau=0.85$ and the maximum number of refinement iterations to $T=3$ for all experiments.

If $C(x,\hat{x}') \geq \tau$, the draft is accepted. Otherwise, the Meta-Agent updates the topology according to the mismatch pattern in Table~\ref{tab:topology_policy}. If no candidate reaches $\tau$ within $T$ iterations, MAS-Bai returns the candidate with the highest cycle-consistency score.

\begin{table}[t]
\centering
\footnotesize
\setlength{\tabcolsep}{3pt}
\begin{tabular}{@{}p{0.38\linewidth}p{0.50\linewidth}@{}}
\toprule
\textbf{Mismatch signal} & \textbf{Topology update} \\
\midrule
Bowen-loan markers missing or altered in $\hat{x}'$
& Activate or re-run $A_{\mathrm{ety}}$ and require explicit phonetic-loan glosses. \\

SOV or postpositional structure not recovered
& Activate or re-run $A_{\mathrm{syn}}$ and require a reordered intermediate representation. \\

Both lexical and syntactic mismatches occur
& Switch to the full quality-oriented pathway:
$A_{\mathrm{ety}} + A_{\mathrm{syn}} + A_{\mathrm{retr}} + A_{\mathrm{trans}}$. \\

Low semantic similarity without lexical/syntactic mismatch
& Refresh retrieved examples and re-run $A_{\mathrm{retr}}$ and $A_{\mathrm{trans}}$. \\

Repeated failure after $T$ iterations
& Return the candidate with the highest $C(x,\hat{x}')$. \\
\bottomrule
\end{tabular}
\caption{Topology modification policy in the self-evolution loop.}
\label{tab:topology_policy}
\end{table}

Cycle-consistency is used as a heuristic reliability signal to detect large semantic or philological drift before final output selection.

\section{Experiments}

We evaluated MAS-Bai in terms of overall translation quality, component contributions, adaptive routing behavior, and cycle-consistency reliability. We further used stratified and human evaluations to examine whether the observed improvements addressed Bai/Bowen-specific translation challenges.

\subsection{Experimental Setup}

We used the Bai--Chinese corpus described in Section~\ref{sec:dataset} and Appendix~\ref{app:corpus_construction}. The 6,536 sentence pairs were partitioned into a 5,000-pair Knowledge Base and a 1,536-sentence test set. The Knowledge Base was used for retrieval, in-context demonstration construction, and fine-tuning the lightweight reverse translator $M_{\mathrm{rev}}$. The test set was reserved exclusively for final evaluation. It contained examples from dialogue-like utterances, formal narrative passages and \textit{Dabenqv} lyric or verse lines.

\begin{table*}[t]
\centering
\small
\setlength{\tabcolsep}{6pt}
\begin{tabular}{llccc}
\toprule
\textbf{Model Backbone} & \textbf{Method} & \textbf{BLEU} & \textbf{ChrF++} & \textbf{COMET} \\
\midrule
\multicolumn{5}{l}{\textit{Baselines: Zero-Shot Performance}} \\
Gemini-1.5-Pro & 0-shot Translation & 0.42 & 10.15 & 22.14 \\
GPT-4o & 0-shot Translation & 0.85 & 12.42 & 26.58 \\
Claude-3.7 & 0-shot Translation & 0.93 & 12.88 & 27.05 \\
Qwen2.5-72B & 0-shot Translation & 0.51 & 10.89 & 23.41 \\
\midrule
\multicolumn{5}{l}{\textit{Baselines: In-Context Learning (Backbone: Qwen2.5-72B)}} \\
Qwen2.5-72B & 3-shot ICL & 1.89 & 14.33 & 35.62 \\
Qwen2.5-72B & 5-shot ICL & 2.14 & 14.89 & 38.17 \\
Qwen2.5-72B & 10-shot ICL & 2.65 & 15.42 & 40.25 \\
\midrule
\multicolumn{5}{l}{\textit{Baselines: Retrieval-Augmented Generation (Backbone: Qwen2.5-72B)}} \\
Qwen2.5-72B & Standard RAG ($k=5$) & 8.12 & 24.56 & 55.30 \\
\midrule
\multicolumn{5}{l}{\textit{Baselines: Forward LoRA (Backbone: Qwen2.5-72B)}} \\
Qwen2.5-72B & Forward LoRA & 13.45 & 31.20 & 57.80 \\
\midrule
\multicolumn{5}{l}{\textit{Ours (Backbone: Qwen2.5-72B)}} \\
\textbf{MAS-Bai} & \textbf{Adaptive orchestration} & \textbf{16.23} & \textbf{35.74} & \textbf{62.15} \\
\bottomrule
\end{tabular}
\caption{
Main results on the Bai--Chinese test set.
Qwen2.5-72B denotes Qwen-2.5-72B-Instruct.
COMET scores are scaled by 100.
}
\label{tab:main_results}
\end{table*}

\paragraph{Baselines.}
We compared MAS-Bai with five baseline families.

\begin{itemize}
    \item \textbf{Zero-shot Translation.} We prompted GPT-4o, Gemini-1.5-Pro, Claude-3.7 and Qwen-2.5-72B-Instruct to translate Bai sentences without demonstrations.

    \item \textbf{Few-shot ICL.} Using Qwen-2.5-72B-Instruct, we evaluated 3-, 5- and 10-shot ICL with randomly selected demonstrations.

    \item \textbf{Standard RAG.} We provided Qwen-2.5-72B-Instruct with the five most similar Bai--Chinese pairs retrieved from the Knowledge Base.

    \item \textbf{Forward LoRA.} We fine-tuned Qwen-2.5-72B-Instruct on the same 5,000 pairs for direct Bai-to-Chinese translation.

    \item \textbf{Static MAS.} We applied the full agent chain
    ($A_{\mathrm{ety}} \to A_{\mathrm{syn}} \to A_{\mathrm{retr}} \to A_{\mathrm{trans}}$) to every input, using the same backbone and agent implementations as MAS-Bai while disabling conditional routing.
\end{itemize}

\paragraph{Implementation Details.}
We employed Qwen-2.5-72B-Instruct \citep{qwen2025qwen25} as the backbone for the Meta-Agent and all functional agents. For the Knowledge Base, we used bge-m3 \citep{chen2024m3embedding} as the embedding model to encode Bai and Chinese sentences. The reverse translator $M_{\mathrm{rev}}$ was a Qwen-2.5-7B model fine-tuned on the same 5{,}000 pairs via LoRA.

In the self-evolution loop, we set the maximum number of refinement iterations to $T=3$ and the acceptance threshold to $\tau=0.85$. These values were fixed before test-set evaluation. Appendix~\ref{app:cycle_sensitivity} examines nearby threshold values and a larger iteration cap. We reported BLEU, ChrF++ and COMET \citep{papineni2002bleu,popovic2017chrf,rei2020comet}. BLEU was computed with SacreBLEU \citep{post2018sacrebleu} using the Chinese tokenizer, and ChrF++ was computed with word-order augmentation. COMET was computed with the same COMET-DA checkpoint for all systems.

The hypothesis and reference were both Modern Chinese, while the source was Bowen-script Bai, which falls outside the source-language distribution typically represented in MT metric training data. We therefore treated COMET as a reference-based measure of target-side semantic similarity and interpreted it alongside human evaluations of adequacy, fluency and Bowen hallucination.

All locally hosted experiments were conducted on six NVIDIA A800 GPUs. The proprietary-model baselines were accessed through their respective APIs.

\subsection{Main Results}
\label{sec:main_results}

Table~\ref{tab:main_results} reports translation performance on the Bai--Chinese test set. Under zero-shot prompting, all four general-purpose LLMs perform poorly: their BLEU scores remain below 1.0, and their COMET scores range from 22.14 to 27.05. Their outputs are often fluent but semantically unrelated, a pattern consistent with the tendency to interpret Bowen characters according to their standard Chinese meanings. Increasing the number of in-context examples improves performance, with 10-shot ICL reaching 2.65 BLEU and 40.25 COMET, but it remains substantially below the retrieval- and adaptation-based systems.

Standard RAG reaches 8.12 BLEU, 24.56 ChrF++, and 55.30 COMET, indicating that retrieved Bai--Chinese examples provide useful translation evidence. Forward LoRA further improves performance to 13.45 BLEU, 31.20 ChrF++, and 57.80 COMET. MAS-Bai achieves the best overall results, with 16.23 BLEU, 35.74 ChrF++, and 62.15 COMET. Compared with Standard RAG, MAS-Bai improves by 8.11 BLEU, 11.18 ChrF++, and 6.85 COMET. Compared with Forward LoRA, it improves by 2.78 BLEU, 4.54 ChrF++, and 4.35 COMET.

These results establish retrieval and supervised adaptation as strong comparison points, while MAS-Bai achieves further improvements. The following sections analyze these gains across system components, linguistic subsets, verifier settings and human judgments.

\subsection{Attribution of Performance Gains}
\label{sec:attribution}

To clarify the sources of MAS-Bai's gains, we examined two complementary factors: Bai-specific philological expertise and adaptive routing. The expert ablations assessed the contribution of language-specific knowledge, while the routing-policy comparison assessed pathway selection with the expert pool and generation settings fixed. Section~\ref{sec:verification_reliability} examines self-evolution.

\paragraph{Contribution of Bai-Specific Experts.}

Table~\ref{tab:ablation} summarizes the component ablations, all of which used Qwen-2.5-72B-Instruct and left the remaining workflow unchanged. Removing the Etymologist reduced BLEU by 4.78 points, from 16.23 to 11.45, consistent with its role in resolving ambiguous Bowen phonetic loans. Removing the Syntactician reduced BLEU by 3.15 points, to 13.08, supporting the value of explicit structural reordering for Bai--Chinese word-order divergence.

\begin{table}[t]
\centering
\small
\setlength{\tabcolsep}{4pt}
\begin{tabular}{lccc}
\toprule
\textbf{Configuration} & \textbf{BLEU} & \textbf{ChrF++} & \textbf{COMET} \\
\midrule
\textbf{MAS-Bai (Full)} & \textbf{16.23} & \textbf{35.74} & \textbf{62.15} \\
\midrule
\multicolumn{4}{l}{\textit{Ablation of Philological Agents}} \\
w/o Etymologist & 11.45 & 28.12 & 51.50 \\
w/o Syntactician & 13.08 & 30.55 & 55.88 \\
\midrule
\multicolumn{4}{l}{\textit{Ablation of Adaptive Routing}} \\
Static MAS & 14.51 & 32.19 & 58.23 \\
\midrule
\multicolumn{4}{l}{\textit{Ablation of Self-Evolution}} \\
w/o Self-Evolution & 12.87 & 30.14 & 54.91 \\
\bottomrule
\end{tabular}
\caption{Ablation study of MAS-Bai components.}
\label{tab:ablation}
\end{table}

These ablations show that both specialists contribute Bai-specific knowledge to translation. Their effects are not additive because the specialists interact with retrieval and final generation.

\paragraph{Contribution of Adaptive Routing.}
To assess the contribution of pathway selection, we kept the backbone, Bai-specific expert pool, prompts, retrieval base, decoding settings and cycle-consistency configuration fixed and changed only the routing policy. We compared MAS-Bai with three routing policies. Uniform random routing sampled the two pathways with equal probability, Static MAS always selected the quality-oriented pathway, and a retrospective COMET oracle selected the higher-scoring pathway for each test sentence according to sentence-level COMET.

\begin{table}[t]
\centering
\footnotesize
\setlength{\tabcolsep}{2pt}
\begin{tabular}{lccc}
\toprule
\textbf{Routing policy} & \textbf{BLEU} & \textbf{COMET} & \textbf{Avg. tokens} \\
\midrule
Uniform random & 14.85 & 59.10 & 1012 \\
Static MAS & 14.51 & 58.23 & 1250 \\
\textbf{MAS-Bai routing} & \textbf{16.23} & \textbf{62.15} & \textbf{775} \\
COMET oracle & 16.65 & 62.80 & -- \\
\bottomrule
\end{tabular}
\caption{
Routing-policy comparison. All variants use the same backbone, expert pool, prompts, retrieval base, decoding settings and cycle-consistency configuration. Avg.\ tokens is the mean cumulative number of input and output tokens across all invoked agents and refinement rounds, counted with the Qwen-2.5 tokenizer. The COMET oracle is retrospective and is included only as a diagnostic upper bound.
}
\label{tab:routing_policy}
\end{table}

MAS-Bai outperforms Static MAS by 1.72 BLEU and 3.92 COMET while using 38\% fewer average tokens. It also exceeds uniform random routing by 1.38 BLEU and 3.05 COMET while using 23\% fewer tokens. MAS-Bai remains close to the retrospective COMET oracle, with gaps of 0.42 BLEU and 0.65 COMET. Since the expert pool and generation settings are shared across these policies, the results indicate an additional contribution from conditional pathway selection.

\subsection{Inference-Time Adaptation}
\label{sec:analysis_adaptation}

The routing-policy comparison in Table~\ref{tab:routing_policy} evaluates the downstream quality and efficiency of pathway selection, while the human-labeled analysis assesses the linguistic plausibility of the routing decisions. We next examined whether the selected pathways corresponded to human judgments of input complexity. On the test set, the Meta-Agent selected the efficiency-oriented pathway for 72\% of simple dialogue-like inputs and the quality-oriented pathway for 85\% of complex \textit{Dabenqv}/Bowen inputs.

In a 200-sentence evaluation, the Meta-Agent achieved 88.5\% accuracy against adjudicated human labels, with Cohen's $\kappa=0.84$ between annotators. This agreement indicates that the rule-informed router largely follows the same complexity distinction as Bai-language researchers. Detailed diagnostics are provided in Appendix~\ref{app:routing_efficiency}.

\subsection{Stratified Results by Register and Linguistic Phenomenon}
\label{sec:analysis_phenomena}

We next examined whether MAS-Bai's improvements held across sentence registers and whether they were concentrated on the Bai/Bowen phenomena targeted by the agent design. Table~\ref{tab:stratified_results} reports COMET scores for the full test set, three register subsets and two diagnostic linguistic-phenomenon subsets. The register analysis covers dialogue-like utterances, formal narrative passages and lyric or verse lines. The Bowen Script Subset contains sentences in which ambiguous phonetic loans are translation-critical, while the SOV Syntax Subset contains sentences requiring substantial reordering into standard Chinese word order.

\begin{table}[t]
\centering
\small
\setlength{\tabcolsep}{2pt}
\begin{tabular}{lrrrr}
\toprule
\textbf{Subset} & \textbf{$n$} & \textbf{Std. RAG} & \textbf{MAS-Bai} & \textbf{$\Delta$} \\
\midrule
Full test set & 1,536 & 55.30 & 62.15 & +6.85 \\
\midrule
\multicolumn{5}{l}{\textit{Register subsets}} \\
Dialogue-like & 512 & 60.15 & 64.20 & +4.05 \\
Formal narrative & 484 & 56.40 & 62.80 & +6.40 \\
Lyric / verse & 540 & 49.75 & 59.60 & +9.85 \\
\midrule
\multicolumn{5}{l}{\textit{Linguistic phenomenon subsets}} \\
Bowen Script Subset & 212 & 48.50 & 73.90 & +25.40 \\
SOV Syntax Subset & 185 & 51.20 & 72.90 & +21.70 \\
\bottomrule
\end{tabular}
\caption{
COMET scores by register and linguistic phenomenon.
COMET scores are scaled by 100.
}
\label{tab:stratified_results}
\end{table}

MAS-Bai outperforms Standard RAG across all three register subsets. The gain increases from 4.05 COMET on dialogue-like inputs to 6.40 on formal narrative passages and 9.85 on lyric or verse lines. This progression indicates that the benefits extend beyond routine utterances and become more pronounced in texts requiring broader contextual or stylistic interpretation.

The largest gains occur on the two diagnostic phenomenon subsets. MAS-Bai improves by 25.40 COMET on the Bowen Script Subset and by 21.70 COMET on the SOV Syntax Subset. These results link the overall performance gains to translation-critical phonetic loans and Bai--Chinese word-order divergence, which are addressed by the Etymologist and Syntactician, respectively.

\subsection{Robustness of Cycle-Consistency Verification}
\label{sec:verification_reliability}

We evaluated the self-evolution mechanism from three perspectives: its contribution to final translation quality, its sensitivity to verifier training data, and its alignment with human adequacy. Removing self-evolution reduced BLEU from 16.23 to 12.87 and COMET from 62.15 to 54.91 (Table~\ref{tab:ablation}). Full verifier comparisons are reported in Appendix~\ref{app:cycle_diagnostics}.

\paragraph{Circularity control.}
We used 4,000 pairs for retrieval and a disjoint set of 1,000 pairs to train $M_{\mathrm{rev\text{-}disjoint}}$. The disjoint verifier achieved 15.88 BLEU and 61.34 COMET, below the shared-verifier setting but above the no-self-evolution variant. The improvement therefore persisted when retrieval and verifier training data did not overlap.

\paragraph{Verifier strength variation.}
With the retrieval base, routing rule, prompts, $\tau=0.85$, and $T=3$ fixed, training the verifier on 250, 1,000, and 5,000 pairs yielded BLEU/COMET scores of 14.12/57.85, 15.65/60.92, and 16.23/62.15, respectively. Even the weakest verifier outperformed the no-self-evolution setting, and performance increased with verifier strength.

\paragraph{Correlation with human adequacy.}
Two bilingual Bai--Chinese researchers rated 150 pre-refinement drafts on a 1--5 adequacy scale. The cycle-consistency score correlated positively with their judgments (Spearman's $\rho=0.68$, $p<0.01$), supporting its use as a heuristic screening signal for low-quality drafts.

These analyses show that self-evolution improves translation quality, remains effective when retrieval and verifier training data are disjoint, and yields stronger results as the reverse translator is trained on more data.

\subsection{Human Evaluation}
\label{sec:human_eval}

We conducted human evaluation to assess translation quality beyond automatic metrics. We randomly sampled 100 sentences from the test set, with proportional coverage of dialogue-like, formal narrative and lyric or verse registers. Three native Bai speakers with linguistic training evaluated anonymized outputs from zero-shot GPT-4o, Standard RAG and MAS-Bai without seeing system identities.

The evaluation covered three dimensions: (1) \textbf{Adequacy}, rated on a 1--5 scale for preservation of source meaning; (2) \textbf{Fluency}, rated on a 1--5 scale for the grammatical correctness and naturalness of the Chinese output; and (3) \textbf{Bowen hallucination rate}, a binary sentence-level label indicating whether a Bai phonetic-loan character was interpreted as its standard Chinese semantic equivalent.

\begin{table}[t]
\centering
\small
\setlength{\tabcolsep}{3pt}
\begin{tabular}{lccc}
\toprule
\textbf{Method} & \textbf{Adeq. $\uparrow$} & \textbf{Flu. $\uparrow$} & \textbf{Halluc. $\downarrow$} \\
\midrule
GPT-4o (0-shot) & 1.84 & 4.12 & 76.0\% \\
Standard RAG & 3.25 & 3.95 & 38.0\% \\
MAS-Bai & \textbf{4.42} & \textbf{4.55} & \textbf{8.0\%} \\
\bottomrule
\end{tabular}
\caption{
Human evaluation on 100 test sentences.
Adequacy and fluency are averaged across three annotators.
}
\label{tab:human_eval}
\end{table}

Table~\ref{tab:human_eval} shows that zero-shot GPT-4o produces fluent Chinese but has low adequacy and a Bowen hallucination rate of 76.0\%. Standard RAG reduces this rate to 38.0\%, while MAS-Bai further reduces it to 8.0\% and achieves the highest adequacy and fluency scores. The 30-percentage-point absolute reduction relative to Standard RAG aligns with the automatic trends and indicates improved source adequacy with fewer Bowen-specific hallucinations.

\section{Discussion}

The results distinguish three sources of improvement. Forward LoRA shows that the Bai--Chinese corpus supports effective supervised adaptation, while MAS-Bai achieves higher overall performance. The expert ablations demonstrate the value of Bai-specific phonetic-loan and syntactic knowledge, and the routing-policy comparison shows an additional benefit from deciding when these capabilities should be invoked. These findings highlight the complementary roles of linguistic specialization and adaptive orchestration.

The phonetic-loan inventory, syntax markers, philological agents and reverse translator constitute the Bai-specific realization of MAS-Bai. Its broader design principle is to identify interpretable linguistic difficulty cues, associate them with specialist capabilities, and use those cues to select and revise an inference pathway. This principle may apply to other shared-glyph writing systems or language pairs with systematic structural divergence. Each application would require its own linguistic analysis, cue inventory, specialist definitions and routing validation. The present evidence establishes the approach within the Bowen-script domains represented in our corpus.

The reliability and efficiency analyses provide a complementary practical perspective. Cycle-consistency serves as an internal screening signal for semantic and philological drift: stronger reverse models yield higher translation scores, and verifier scores correlate with human adequacy. Adaptive routing reduces average token consumption by 38\% relative to Static MAS, partly offsetting the additional cost of iterative verification when input difficulty varies.

\section{Conclusion}

We presented \textbf{MAS-Bai}, a linguistically informed adaptive multi-agent framework, together with a 6,536-sentence Bai--Chinese corpus. On a 1,536-sentence test set, MAS-Bai outperforms zero-shot, ICL, Standard RAG, Forward LoRA and Static MAS baselines. Expert ablations and controlled routing comparisons indicate complementary contributions from Bai-specific philological knowledge and conditional pathway selection, with the largest gains on phonetic-loan ambiguity and divergent Bai word order. Human evaluation and verifier analyses are consistent with the translation results. These findings demonstrate the effectiveness of MAS-Bai within the Bowen-script domains represented in our corpus. Evaluating its transfer to other low-resource languages or shared-glyph writing systems remains future work.

\section*{Limitations}

MAS-Bai was developed and evaluated within the Bowen-script domains represented in our corpus. Its phonetic-loan inventory, specialist definitions and routing cues are Bai-specific, and applying the framework to another language would require new linguistic analysis and routing validation. Although Appendix~\ref{app:retrieval_overlap} finds limited near-duplicate overlap with the Knowledge Base, the sentence-level split does not provide a source- or document-held-out evaluation.

Cycle-consistency is a heuristic reliability signal whose performance depends on the coverage and biases of $M_{\mathrm{rev}}$, and the human evaluation is limited to 100 sentences and three annotators. Self-evolution also introduces additional LLM calls, although adaptive routing reduces unnecessary expert computation.

\section*{Ethics Considerations}

This study draws on an officially published \textit{Dabenqv} collection and a community-held handwritten Bowen-script manuscript, both used with permission from the relevant rights holders or custodians. Native Bai speakers contributed to the transcription and translation of the manuscript materials, and a Bai-language researcher reviewed the resulting translations for adequacy and consistency with the source.

GPT-4o was evaluated as a zero-shot translation baseline. ChatGPT-4o was used separately and solely for clerical assistance during sentence alignment, including standardizing punctuation, matching line breaks, and flagging obvious alignment anomalies. It was not used to judge semantic equivalence, decide sentence-pair inclusion, or filter difficult examples. Human evaluation was conducted by native Bai speakers with linguistic training on anonymized system outputs and did not require annotators to provide personal or sensitive information.

The corpus and annotations are used for research and language-documentation purposes under the permissions governing the source materials. Community-held manuscript content will not be released under an unrestricted open license. Where permissions allow, we will release prompts, configuration files, evaluation scripts, split metadata, annotation schemas, and redacted or derived examples. Requests for access to restricted materials may be considered on a case-by-case basis, subject to approval from the relevant rights holders or custodians. Any access or reuse should comply with the original permissions and respect the cultural context of the materials.

\section*{Acknowledgments}

This work was supported by the National Social Science Fund of China (Grant No.\ 21BYY185). We thank the anonymous reviewers for their insightful comments.

\bibliography{custom}

\clearpage
\appendix

\setcounter{figure}{0}
\renewcommand{\thefigure}{A\arabic{figure}}
\setcounter{table}{0}
\renewcommand{\thetable}{A\arabic{table}}

\begin{table*}[t]
\centering
\small
\setlength{\tabcolsep}{5pt}
\begin{tabular}{@{}p{0.29\textwidth}rrp{0.44\textwidth}@{}}
\toprule
\textbf{Source} & \textbf{Units} & \textbf{Tokens} & \textbf{Role} \\
\midrule
Published \textit{Dabenqv} collection
& 4,986 & $\sim$75K
& Existing Bai--Chinese sentence pairs \\

Handwritten Bowen manuscript
& 1,552 & $\sim$12K
& Bai sentences translated into Chinese \\
\midrule
Candidate units
& 6,538 & $\sim$87K
& Before final alignment and quality control \\

Final aligned corpus
& 6,536 & --
& After sentence-level quality control \\
\bottomrule
\end{tabular}
\caption{Corpus source statistics. Token counts are source-side estimates recorded during data collection.}
\label{tab:corpus_source_stats}
\end{table*}

\section{Detailed Corpus Construction and Token Statistics}
\label{app:corpus_construction}

\paragraph{Source collections.}
The corpus is based on the two Bowen-script sources introduced in Section~\ref{sec:dataset}: an officially published \textit{Dabenqv} collection and an authorized community-held handwritten manuscript. The published source contributes existing Bai--Chinese sentence pairs, while the manuscript source contributes Bai sentence units that were subsequently translated into Modern Chinese. Table~\ref{tab:corpus_source_stats} reports the source-level counts and token estimates.

\paragraph{Digitization.}
We digitized the two sources using source-appropriate workflows. For the printed \textit{Dabenqv} collection, we scanned the published pages and applied OCR to recognize Bai characters, producing an initial electronic transcription. The OCR outputs were then manually checked to correct recognition errors, remove corrupted fragments and standardize sentence boundaries. For the handwritten manuscript, whose orthography and handwriting style are more variable, we produced the electronic transcription primarily by manual typing while preserving the original character forms where possible. As part of the broader documentation workflow, manuscript pages were also scanned or photocopied, and owners' recitations were recorded to preserve original readings and support later annotation.

\paragraph{Translation.}
The candidate sentences from the handwritten source were translated into Modern Chinese by native Bai speakers, including local Bai teachers and college students. A Bai-language researcher then reviewed the Chinese translations for fluency, adequacy and consistency with the original Bowen-script source. This review step was especially important for community manuscript materials, where orthographic variation can make it difficult to distinguish genuine character variants from scribal errors.

\paragraph{Sentence alignment and quality control.}
After digitization and translation, all materials were organized into sentence-level parallel files and aligned at the sentence level. ChatGPT-4o was used only to standardize punctuation, match line breaks and flag obvious length or ordering anomalies. All sentence-pair inclusion and semantic adequacy decisions were made by human annotators or Bai-language researchers.

\section{Retrieval Overlap Diagnostics}
\label{app:retrieval_overlap}

Because both Standard RAG and MAS-Bai retrieve examples from the 5,000-pair Knowledge Base, we examined near-duplicate overlap between the test set and the retrieval pool. Using the same bge-m3 representation and similarity function as the retrieval module, we computed the maximum Knowledge Base similarity for each test sentence. The mean and median maximum similarities were 0.68 and 0.65, respectively, and 18 of 1,536 test sentences (1.2\%) exceeded a similarity threshold of 0.9. These results indicate limited near-duplicate overlap.

\section{Routing and Efficiency Diagnostics}
\label{app:routing_efficiency}

This appendix expands the inference-time adaptation analysis in Section~\ref{sec:analysis_adaptation}. It reports pathway selection for simple and complex inputs, agreement with human complexity labels, and the relation between token consumption and translation quality.

\subsection{Pathway Selection Analysis}

We partitioned the test set into simple and complex subsets using human-informed linguistic criteria. Simple inputs were dialogue-like sentences with transparent script usage and word order requiring limited restructuring. Complex inputs contained dense ambiguous Bowen phonetic loans, \textit{Dabenqv}-style phrasing, Bai word-order patterns that diverge from standard Chinese, or cases requiring explicit structural reordering. We then recorded the pathway selected for each sentence.

\begin{figure}[t]
    \centering
    \includegraphics[width=\linewidth]{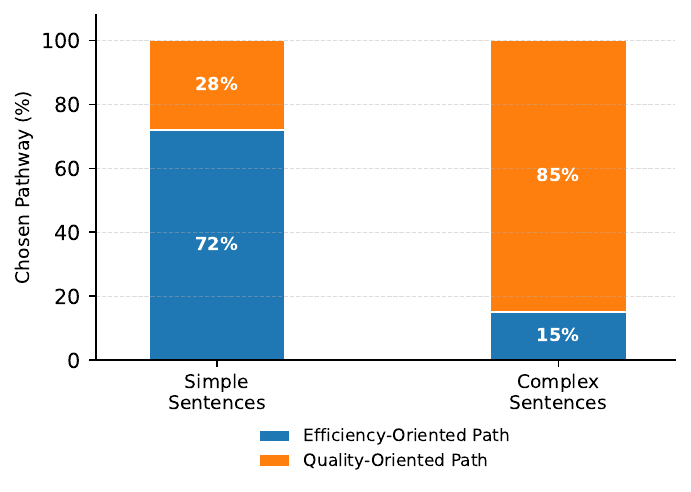}
    \caption{
    Distribution of computational pathways chosen by the Meta-Agent for simple and complex inputs.
    }
    \label{fig:path_distribution_app}
\end{figure}

As shown in Figure~\ref{fig:path_distribution_app}, the Meta-Agent selects the efficiency-oriented pathway for 72\% of simple inputs and the quality-oriented pathway for 85\% of complex inputs. The deeper philological pathway is therefore concentrated on inputs with stronger Bai/Bowen linguistic cues.

\paragraph{Human-labeled routing evaluation.}
We randomly sampled 200 test sentences and asked two Bai-language researchers to independently assign the same \textsc{Simple}/\textsc{Complex} categories. Inter-annotator agreement was strong (Cohen's $\kappa=0.84$), and disagreements were resolved through discussion to form an adjudicated reference set.

\begin{table}[t]
\centering
\small
\setlength{\tabcolsep}{1mm}
\begin{tabular}{lcccc}
\toprule
\textbf{Pathway Class} & \textbf{Precision} & \textbf{Recall} & \textbf{F1} & \textbf{Support} \\
\midrule
Simple (Efficiency)    & 87.6 & 90.2 & 88.9 & 102 \\
Complex (Quality)      & 89.5 & 86.7 & 88.1 & 98  \\
\midrule
\textbf{Overall Accuracy} & \multicolumn{3}{c}{\textbf{88.5}} & 200 \\
\bottomrule
\end{tabular}
\caption{
Meta-Agent routing accuracy against human expert annotations. Metrics are percentages.
}
\label{tab:routing_eval_app}
\end{table}

Table~\ref{tab:routing_eval_app} shows balanced performance across both classes, with 88.5\% overall accuracy, 86.7\% recall for complex inputs and 90.2\% recall for simple inputs. The class-level results support the alignment between the routing rule and human judgments of linguistic complexity.

\subsection{Efficiency and Quality Trade-off}

We compared MAS-Bai with Static MAS, which always executes the full expert chain. Because the two systems share the same backbone and expert implementations, the comparison assesses how conditional routing changes computational cost and translation quality.

\begin{figure}[t]
    \centering
    \includegraphics[width=\linewidth]{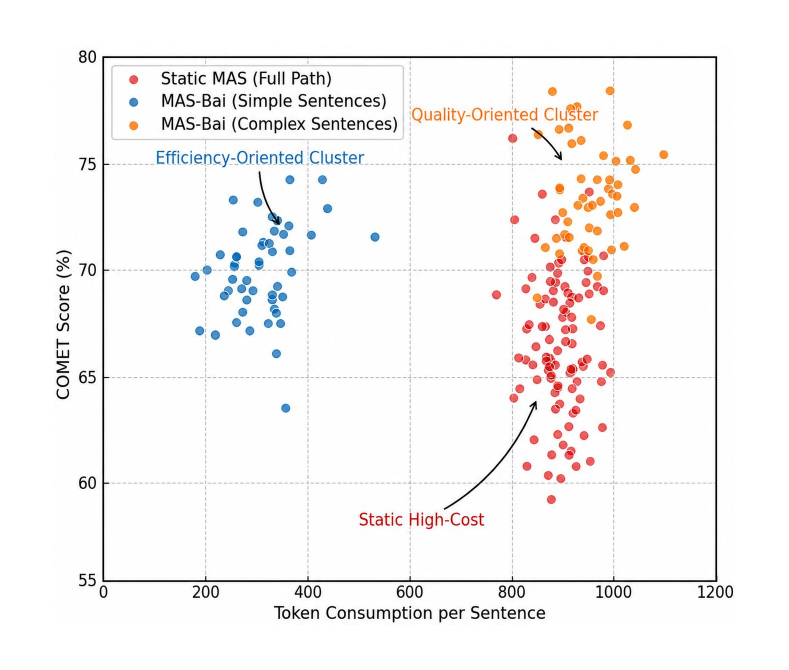}
    \caption{
    COMET score versus token consumption for adaptive MAS-Bai and Static MAS.
    }
    \label{fig:tradeoff_app}
\end{figure}

Figure~\ref{fig:tradeoff_app} plots sentence-level COMET against token consumption. Static MAS occupies a high-token region because every sentence passes through the full expert chain. MAS-Bai shows two operating regions: simple inputs use fewer tokens, while complex inputs retain access to the deeper philological pathway. On average, MAS-Bai reduces inference token consumption by approximately 38\% relative to Static MAS while achieving higher overall translation quality.

\section{Cycle-Consistency Diagnostics}
\label{app:cycle_diagnostics}

\subsection{Verifier Separation and Strength}

Table~\ref{tab:mrev_robustness} reports the complete verifier comparisons underlying Section~\ref{sec:verification_reliability}. The disjoint setting used 4,000 pairs for retrieval and a non-overlapping set of 1,000 pairs for reverse-model training. The verifier-strength settings used the same 5,000-pair retrieval base and differed only in the amount of data used to train $M_{\mathrm{rev}}$.

\begin{table}[t]
\centering
\small
\setlength{\tabcolsep}{3pt}
\begin{tabular}{lccc}
\toprule
\textbf{Setting} & \textbf{Training pairs} & \textbf{BLEU} & \textbf{COMET} \\
\midrule
\multicolumn{4}{l}{\textit{Circularity control}} \\
No self-evolution & -- & 12.87 & 54.91 \\
Shared verifier & 5,000 & 16.23 & 62.15 \\
Disjoint verifier & 1,000 & 15.88 & 61.34 \\
\midrule
\multicolumn{4}{l}{\textit{Verifier strength variation, fixed 5,000-pair retrieval base}} \\
Weak verifier & 250 & 14.12 & 57.85 \\
Medium verifier & 1,000 & 15.65 & 60.92 \\
Full verifier & 5,000 & 16.23 & 62.15 \\
\bottomrule
\end{tabular}
\caption{
Cycle-consistency reliability under verifier-data separation and verifier-strength variation.
}
\label{tab:mrev_robustness}
\end{table}

\subsection{Threshold and Iteration Sensitivity}
\label{app:cycle_sensitivity}

We evaluated nearby acceptance thresholds with $T=3$ and compared the selected setting with a larger iteration cap, while holding the backbone, retrieval base, prompts and reverse verifier fixed. Table~\ref{tab:cycle_sensitivity} reports the results.

\begin{table}[t]
\centering
\small
\setlength{\tabcolsep}{5pt}
\begin{tabular}{cccc}
\toprule
$\tau$ & $T$ & \textbf{BLEU} & \textbf{COMET} \\
\midrule
0.80 & 3 & 15.98 & 61.85 \\
0.85 & 3 & 16.23 & 62.15 \\
0.90 & 3 & 16.15 & 61.95 \\
0.85 & 5 & 17.27 & 62.18 \\
\bottomrule
\end{tabular}
\caption{Sensitivity to the cycle-consistency acceptance threshold and iteration cap.}
\label{tab:cycle_sensitivity}
\end{table}

Performance was stable for nearby values of $\tau$ when $T=3$. Increasing the cap to $T=5$ improved BLEU, while COMET changed little. We retained the pre-specified $T=3$ setting because it limits the maximum number of refinement calls.

\end{document}